\documentclass[letterpaper, 10 pt, conference]{ieeeconf}  

\IEEEoverridecommandlockouts                              

\usepackage{graphicx}
\usepackage{subcaption}
\usepackage[font=small,labelfont=bf]{caption}
\usepackage{mwe}
\usepackage{amsmath}
\usepackage{tablefootnote}
\usepackage{url}
\usepackage{listings}
\usepackage{float}
\usepackage{multirow}
\usepackage{hhline} 
\usepackage{tablefootnote}
\usepackage{dingbat}
 \usepackage[table]{xcolor}
\usepackage[dvipsnames]{xcolor}
\usepackage{tikz}
\usepackage{bbding}
\usepackage{color}
 \usepackage[table]{xcolor}
\usepackage{cuted}
\usepackage{capt-of}
\usepackage{makecell}
\usepackage{tabularx}
\definecolor{cvprblue}{rgb}{0.21,0.49,0.74}
\usepackage{arydshln}
\usepackage[pagebackref,breaklinks,colorlinks,citecolor=cvprblue]
{hyperref}
\usepackage{array} 
\usepackage{booktabs} 
\usepackage{amsmath}  
\usepackage{threeparttable} 
\usepackage{pifont}
\usepackage{xcolor} 
\newcommand{\xmark}{{\color{red}\ding{55}}} 
\usepackage[table]{xcolor}
\newcolumntype{Y}{>{\raggedright\arraybackslash}X}
\title{\LARGE \bf
\textsc{PedestrianQA}: A Benchmark for Vision-Language Models on Pedestrian Intention and Trajectory Prediction
}

\author{
    Naman Mishra, Shankar Gangisetty, C.V. Jawahar%
    \thanks{CVIT, IIIT-Hyderabad, India.
    {\tt\small \{naman.mishra@research., shankar.gangisetty@ihub-data., jawahar@\}iiit.ac.in}}%
}

\begin{document}
\maketitle

\begin{abstract}
Pedestrian intention and trajectory prediction are critical for the safe deployment of autonomous driving systems, directly influencing navigation decisions in complex traffic environments. Recent advances in large vision–language models offer a powerful new paradigm for these tasks by combining high-capacity visual understanding with flexible natural language reasoning. In this work, we introduce PedestrianQA, a large-scale video-based dataset that formulates pedestrian intention and trajectory prediction as question–answering tasks augmented with structured rationales. PedestrianQA expresses richly annotated pedestrian sequences, in natural language, enabling VLMs to learn from visual dynamics, contextual cues, and interactions among traffic agents while generating concise explanations of their predictions without needing specialized architectures tailored for each task. Empirical evaluations across PIE, JAAD, TITAN, and IDD-PeD show that finetuning state-of-the-art VLMs on PedestrianQA significantly improves intention classification, trajectory forecasting accuracy, and the quality of explanatory rationales, demonstrating the strong potential of VLMs as a unified and explainable framework for safety-critical pedestrian behavior modeling. Dataset and models are available at \url{https://github.com/botmahn/PedestrianQA}
\end{abstract}

\section{Introduction}
\label{Sec_Introduction}
\noindent Ensuring pedestrian safety remains one of the most critical challenges in deploying autonomous vehicles (AVs). Pedestrians are highly unpredictable, often exhibiting complex behaviors shaped by dynamic environments and interactions with other agents. Accurately anticipating whether a pedestrian intends to cross and forecasting their future trajectory are essential for safe navigation in structured (e.g., signalized crosswalks) and unstructured settings (e.g., dense urban roads without explicit rules). Failures in these capabilities have repeatedly been linked to AV disengagements and hazardous incidents.

\begin{figure}[t]
  \centering
  \includegraphics[width=\columnwidth]{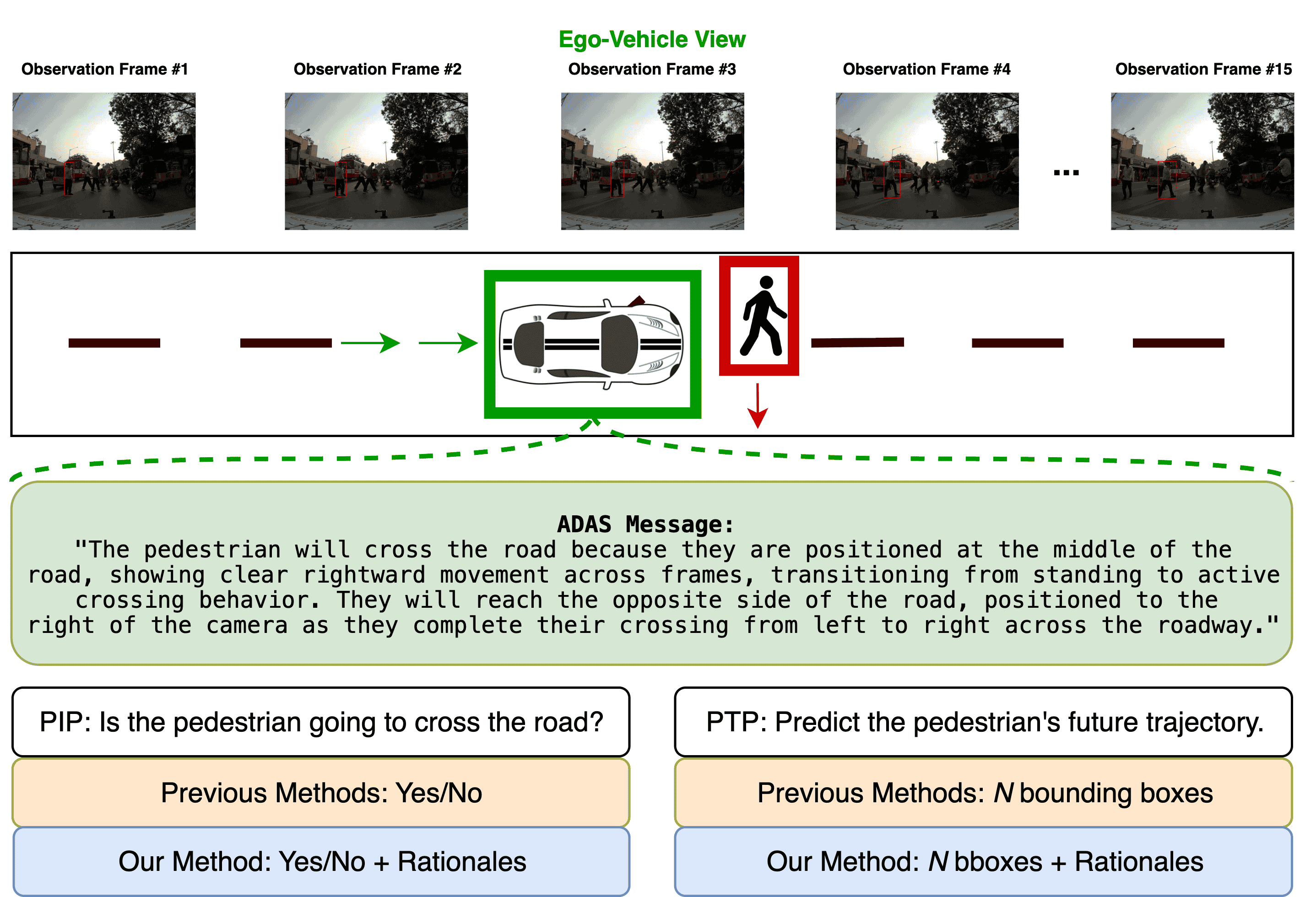}
  \caption{An illustration of an unstructured-traffic scenario where a pedestrian stands in the middle of the road in front of the ego-vehicle, attempting to cross the road. Unlike prior approaches that provide only predictions, our method predicts the intention and trajectory and generates supporting rationales.}
  \label{fig:teaser_diagram}
\end{figure}

The risk is amplified in unstructured environments~\cite{idd2025ped}, where pedestrian movement and behavior are especially erratic. Consider as illustrated in Fig.~\ref{fig:teaser_diagram}, where a pedestrian may attempt to cross a road segment lacking marked crosswalks and traffic lights, suddenly stepping from between parked vehicles without signaling or checking for oncoming traffic, forcing the ego-vehicle to brake abruptly to avoid collision. Motivated by such scenarios, we aim to advance Advanced Driver Assistance Systems (ADAS) with natural-language explanations that enhance explainability in both structured and unstructured traffic scenarios, build user trust, and support scalable deployment.

Within this context, we focus on two fundamental tasks: Pedestrian Intention Prediction (PIP)~\cite{JAAD, PIE, idd2025ped, PIE++, pedVLM}, which predicts whether a pedestrian \emph{intends} to cross, and Pedestrian Trajectory Prediction (PTP)~\cite{TITAN, LGTraj, PIE, JAAD, idd2025ped}, which forecasts the pedestrian’s spatio-temporal path. Both are short-horizon prediction problems that demand rapid responses from traffic agents to prevent accidents. Despite steady progress, existing methods remain limited in two key respects. First, they underutilize multimodal reasoning, failing to integrate the rich semantics of traffic context, agent interactions, and pedestrian attributes within a unified framework. Second, they treat prediction as a black-box classification or regression problem, offering little to no explanation of their predictions, and thereby hindering user trust and large-scale deployment. 

\begin{figure*}[t]
\centering
\includegraphics[width=\textwidth]{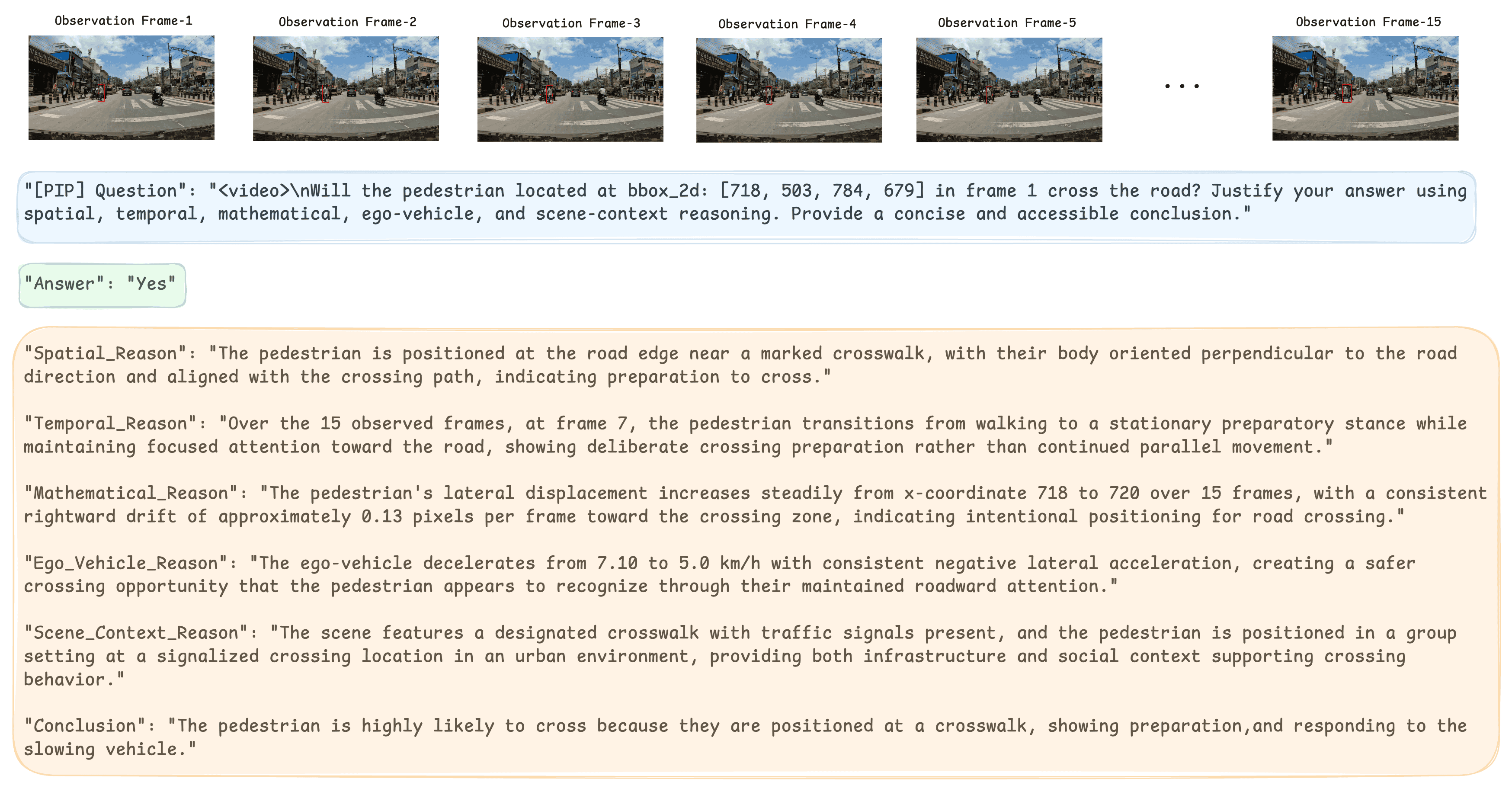}
\caption{\textbf{PedestrianQA Dataset PIP Sample.} The top row shows the observation frames. The ``Question" and ``Answer" are followed by 5 types of rationales and a conclusion.}
\label{fig:pip_example}
\end{figure*}

To address these limitations, we explore large-scale vision–language models (VLMs)~\cite{instructblip,qwen2.5vl,internvl3,llava,gemma3} that have revolutionized the potential use-cases of fusing vision and language modalities. Trained on internet-scale corpora of paired images, videos, and text, they perform strongly in tasks such as visual question answering (VQA), generating detailed scene description, optical character recognition (OCR), and object-grounded spatio-temporal reasoning. With the advent of methods that increase large models efficiency~\cite{dettmers2022llmint8, dao2022flashattention}, VLMs have successfully been applied to several real-time tasks~\cite{vlm_edge}, including autonomous driving~\cite{ad_survey,jiang2025alphadrive}. Yet, their application to pedestrian-level safety-critical tasks like PIP and PTP remains limited. Existing pedestrian benchmarks~\cite{PIE,JAAD,TITAN,idd2025ped} provide rich spatial–temporal data, behavioral categories, environmental context, and ego-vehicle signals that can be reformulated in natural language. Integrating these textual representations with visual inputs can enable VLMs to model pedestrian behavior comprehensively and in a unified manner.

To this end, we introduce \textbf{PedestrianQA}, a multimodal video-based question–answering dataset specifically designed for pedestrian intention and trajectory prediction. Unlike prior datasets, PedestrianQA frames pedestrian behavior understanding as a video question–answering–explanation task, enabling VLMs to predict crossing intention and forecast future trajectories while simultaneously generating explanatory natural-language rationales for their decisions. Each pedestrian sequence is associated with a question, an answer, and five structured rationales that capture complementary aspects of the scene. These include spatial information (e.g., body pose, positioning, stride), temporal motion cues (e.g., acceleration, sudden stops, changes in walking speed), scene-level context (e.g., presence of crosswalk markings, traffic signals, traffic density, visibility conditions), and ego-vehicle interactions (e.g., decelerating, yielding, or assertive). For example, a rationale might describe a pedestrian leaning forward with an extended stride (spatial), increasing walking pace over the past two seconds (temporal), approaching an unsignalized intersection with moderate vehicle traffic (scene context), while the ego-vehicle slows and flashes hazard lights (interaction), collectively supporting the prediction that the pedestrian will cross imminently. Overall, this design requires models not only to predict \emph{what} will happen (intention and trajectory) but also to explain \emph{why} those outcomes are plausible. One such sample is depicted in Fig.~\ref{fig:pip_example}. Further, we show that state-of-the-art VLMs finetuned on PedestrianQA achieve substantial gains in predictive accuracy and reasoning quality. Finally, PedestrianQA enables evaluation metrics that jointly assess prediction correctness and explanation quality, advancing transparent, safety-critical decision-making in autonomous driving.

In summary, our contributions are: \begin{itemize}\item \textbf{PedestrianQA}: a novel multimodal dataset that unifies pedestrian intention and trajectory prediction with fine-grained, rationale-based question–answer annotations.\item \textbf{Baseline}: a strong prediction and reasoning baseline established by finetuning a state-of-the-art VLM on PedestrianQA, demonstrating how multimodal models can be adapted to pedestrian intention and trajectory prediction.\item \textbf{Benchmark}: comprehensive experiments showing that VLMs finetuned on PedestrianQA outperform existing baselines in intention and trajectory prediction, as well as rationale quality.\end{itemize}

\section{Related Works}
\label{realted-works}
\subsection{Pedestrian Intention and Trajectory Prediction}

Several datasets~\cite{PIE,JAAD,TITAN,idd2025ped,PIE++,pedVLM,PIPNet, STIP, pedx} support this line of work. JAAD~\cite{JAAD} pioneered PIP using behavioral and contextual cues but lacked scale, ego-vehicle data, and interacting vehicles localization. PIE~\cite{PIE} addresses these gaps by introducing vehicle trajectories and ego-vehicle odometry, while TITAN~\cite{TITAN} offers fine-grained action priors and detailed scene dynamics. Finally, IDD-PeD~\cite{idd2025ped} targets unstructured traffic scenes with expanded pedestrian behavioral attributes. We refer readers to Table~\ref{tab:dataset-annotations} for details. While these datasets offer diverse information, they lack inherent integration or shared structure. Further, they do not provide explicit rationales capturing these underlying relationships.

To address this gap, PIE++~\cite{PIE++} augments PIE with manually annotated sentence-level rationales, and PedVLM~\cite{pedVLM} introduces PedPrompt, a binary classification question–answering corpus. These datasets have drawbacks in that PIE++ remains limited in scale and annotation diversity, and PedPrompt reduces intention to binary labels, overlooking the richer information in the TRANS~\cite{TRANS} dataset and failing to unify it into a descriptive natural language resource. In contrast, our approach automatically generates questions and answers along with diverse, fine-grained, object-grounded natural language rationales that integrate cues from all scene elements for explainable pedestrian behavior prediction.

Parallely, several video question–answer datasets~\cite{DriveLM,driveVLM,mm_au,roadsocial,rank2tell,nuscenes_qa, LingoQA, drama, lpad} exist in the domain of autonomous driving. We highlight RoadSocial~\cite{roadsocial}, and Rank2Tell~\cite{rank2tell} due to similarities with our own work. RoadSocial leverages multi-view videos sourced from \texttt{x.com} and generates a large-scale question-answer-reasoning dataset using a state-of-the-art LLM~\cite{claude_sonnet_4}. Rank2Tell provides multimodal urban-intersection videos with synchronized RGB, LiDAR, and CAN data, annotated for important-agent localization, importance ranking, and natural language explanations. In contrast, our work is explicitly pedestrian-centric, emphasizing predictions with object-grounded spatio-temporal reasoning.

\noindent \textbf{Methodologically:} Early approaches fused visual cues with SVMs~\cite{JAAD}, LSTM encoders~\cite{PIE}, or I3D-based action models combined with ego-motion~\cite{TITAN}. In the line of spatio-temporal fusion, PCPA~\cite{eval_paper}, MaskPCPA~\cite{MaskPCPA}, and PIP-Net~\cite{PIPNet} apply temporal attention over multimodal streams. In traditional vision-language methods, PIE++~\cite{PIE++} employs MINDREAD, a cross-modal encoder that fuses visual features with textual rationales. PedVLM~\cite{pedVLM} combines visual tokens via CLIP~\cite{clip} and text tokens, for generating binary crossing intention labels from T5~\cite{T5}. Rank2Tell integrates 2D CNNs and 3D point clouds in a relational graph for joint ranking and captioning. Models such as LG-Traj~\cite{LGTraj} and IntentFormer~\cite{intentformer} jointly encode visual and trajectory cues, and ClipCross~\cite{ClipCross} aligns CLIP image–text embeddings for intention classification. Utilizing large VLMs, among recent work~\cite{GPT4VWheel, pip_vlm}, GPT-4V~\cite{GPT4VWheel} and VLMs guided by hierarchical prompts~\cite{pip_vlm} explore zero-shot performances on PIP. In contrast to these methods, we use a single open-source VLM for intention prediction, trajectory prediction, and rationale generation, all without any architectural modifications.

\subsection{Vision-Language Models}

Early VLMs extended pre-trained vision encoders with instruction-tuned LLMs, for example, InstructBLIP~\cite{instructblip} builds on BLIP-2~\cite{blip2} with instruction tuning on open-source datasets, while LLaVA~\cite{llava} combines vision transformers~\cite{vit, clip} with the Vicuna~\cite{vicuna2023} for conversational tasks.  
\newline
Recent models~\cite{llavanextvideo, qwen2.5vl, internvl3, kwai-keye} can scale to multi-image and long-video inputs. LLaVA-NeXT~\cite{llavanextvideo} generalizes to videos using AnyRes frame packing, linear RoPE~\cite{RoPE} scaling for long sequences, and video supervised finetuning. Qwen2.5-VL~\cite{qwen2.5vl} introduces a novel dynamic-resolution Vision Transformer and absolute time encoding for fine-grained grounding, and long-video comprehension. InternVL3~\cite{internvl3} employs native multimodal pre-training with variable visual position encoding, advanced post-training, and test-time scaling. Building on advances in reasoning for LLMs, recent work adds explicit reasoning capability to VLMs~\cite{mmreasoning_survey}. Kwai Keye-VL~\cite{kwai-keye}, for instance, targets short-form video with a multi-stage training recipe and a dedicated \emph{thinking} mode for enhanced reasoning.

Several VLM methods have also been adapted to autonomous driving. Dolphins~\cite{dolphins} adapts OpenFlamingo~\cite{openflamingo} with Grounded Chain-of-Thought (CoT) reasoning and driving-specific instruction tuning. OmniDrive~\cite{omnidrive} builds on LLaVA with two complementary designs: Omni-L and Omni-Q for joint vision-language alignment and driving-scene reasoning. DriveLM~\cite{DriveLM} extends BLIP-2 by chaining perception, prediction, and planning QAs in a graph, using context from preceding nodes to reason about the scene and generate a natural-language behavior description. In contrast, we prioritize fine-grained modeling of pedestrian dynamics and spatio-temporally grounding traffic agents to predict intentions, forecast trajectories, and provide explanatory rationales, all without making any architectural changes.

\begin{figure*}[t]
\centering
\includegraphics[width=\textwidth]{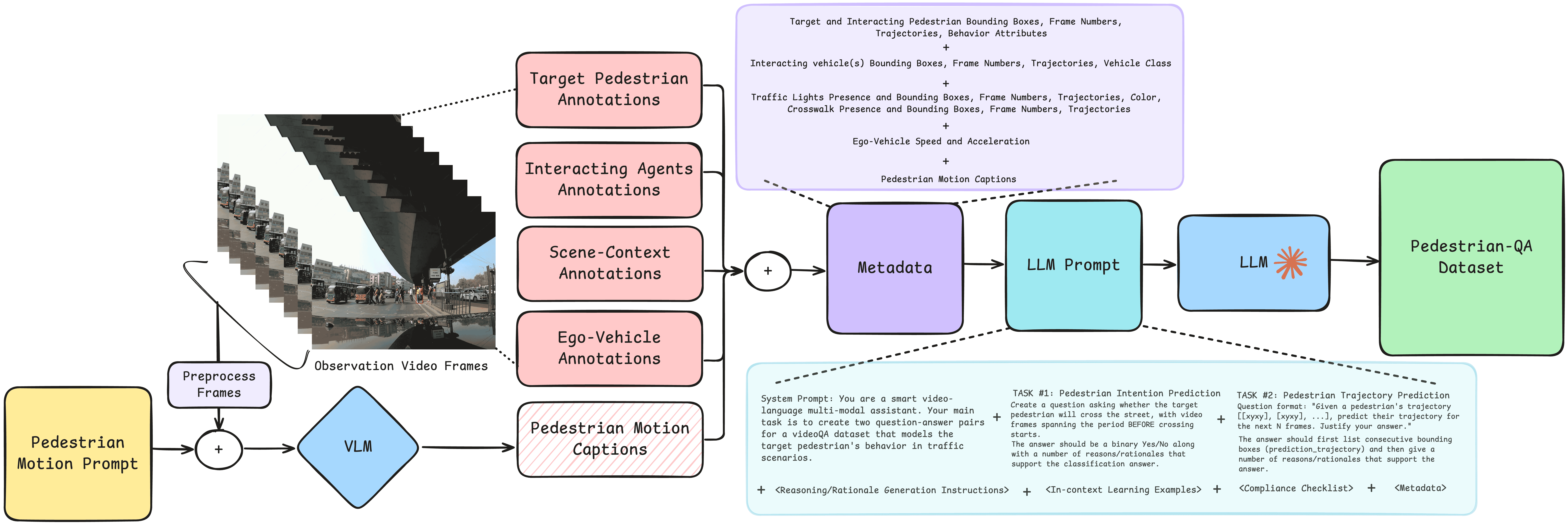}
\caption{\textbf{Data generation pipeline}: We first aggregate all ground-truth, human-annotated annotations from the constituent datasets into a unified metadata schema. We use generated VLM captions to enrich motion semantics using carefully designed pedestrian-motion prompts that target fine-grained cues. These captions are validated for format and appended to the metadata. We then construct a single instruction package containing: (i) a system prompt, (ii) task definitions for PIP and PTP, (iii) step-by-step guidance for producing structured, fine-grained rationales, (iv) a small set of in-context exemplars, (v) a compliance checklist for high-quality rationale generation, and (vi) the sequence-level metadata tables. This package is provided to \texttt{claude-sonnet-4-20250514} LLM-API to generate triplets of questions, answers, and rationales.}
\label{fig:pipeline}
\end{figure*}
\section{PedestrianQA Dataset}
In this section, we describe the generation process of our PedestrianQA dataset, with the full pipeline illustrated in Fig.~\ref{fig:pipeline} and a PIP example of our dataset in Fig.~\ref{fig:pip_example}
\subsection{Data Collection}
We construct our PedestrianQA corpus from four publicly available pedestrian datasets: 3 structured (JAAD~\cite{JAAD}, PIE~\cite{PIE}, TITAN~\cite{TITAN}), and 1 unstructured (IDD-PeD~\cite{idd2025ped}). We selected these datasets for their comprehensive annotations, including localization across frames for target pedestrians, interacting pedestrians and vehicles, and traffic control elements like crosswalks and traffic lights, detailed labels of pedestrian and ego-vehicle behaviors. We refer readers to Table~\ref{tab:dataset-annotations} for details. Each instance in our dataset features a unique target pedestrian observed from the ego-vehicle, along with surrounding interacting pedestrians, non-ego interacting vehicles, and traffic-control elements such as crosswalks and traffic lights. In subsequent sequences, the target pedestrian from an earlier sequence can act as an interacting pedestrian.

\subsection{Sequence Sampling}
Following~\cite{eval_paper}, we sample $0.5$\,s observational frames from JAAD~\cite{JAAD}, PIE~\cite{PIE}, and IDD-PeD~\cite{idd2025ped}, each of which is recorded at approximately $30$ fps. A Time-To-Event (TTE) between $1–2$\,s is defined such that the final frame of each target pedestrian sequence falls within this range. For example, if the annotated crossing point occurs at frame $100$, the sequence ends between frames $40$ and $70$ for a video recorded at $30$ fps. For trajectory prediction in these datasets, we retain only those sequences that contain at least $1.5$\,s of video after the observation period. In TITAN~\cite{TITAN}, we extract observation sequences guided by the \texttt{"Simple Context"} annotation. Specifically, we sample frames with the TTE constraint preceding a change in context to either \texttt{"crossing a street at pedestrian crossing"} or \texttt{"jaywalking (illegally crossing NOT at pedestrian crossing)"}. Furthermore, we leverage the attribute \texttt{"waiting to cross street"} as an additional indicator of positive crossing intention. Following details in~\cite{TITAN}, TITAN is sampled at $10$\,fps, with an observation window of  $1$\,s and prediction up to $2$\,s beyond it. We adopt the same TTE sampling protocol for TITAN, as we did with JAAD~\cite{JAAD}, PIE~\cite{PIE}, and IDD-PeD~\cite{idd2025ped}.

\label{data-capture}

\begin{table}[t]
\centering
\scriptsize
\begin{threeparttable}
\caption{Annotation availability across datasets. ``\checkmark'' indicates availability of the corresponding annotation type.}
\label{tab:dataset-annotations}
\begingroup
\setlength{\tabcolsep}{4pt} 
\renewcommand\arraystretch{1.1} 
\begin{tabularx}{\linewidth}{YYcccc}
\toprule
\makecell[l]{Category} & \makecell[l]{Annotation} & \makecell{IDD-PeD\\\cite{idd2025ped}} & \makecell{JAAD\\\cite{JAAD}} & \makecell{PIE\\\cite{PIE}} & \makecell{TITAN\\\cite{TITAN}} \\
\midrule
\multirow{2}{*}{Target Pedestrian}
 & B.Boxes    & \checkmark & \checkmark & \checkmark & \checkmark \\
 & Trajectory & \checkmark & \checkmark & \checkmark & \checkmark \\
\midrule
\multirow{2}{*}{Interacting Pedestrians}
 & B.Boxes    & \checkmark & \checkmark & \checkmark & \checkmark \\
 & Trajectory & \checkmark & \checkmark & \checkmark & \checkmark \\
\midrule
Pedestrian Behavior & -- & \checkmark & \checkmark & \checkmark & \checkmark \\
\midrule
\multirow{3}{*}{Interacting Vehicles}
 & B.Boxes    & \checkmark & \xmark     & \checkmark & \checkmark \\
 & Trajectory & \checkmark & \xmark     & \checkmark & \checkmark \\
 & Class      & \checkmark & \xmark     & \checkmark & \checkmark \\
\midrule
\multirow{2}{*}{Ego Vehicle}
 & Speed        & \checkmark & \xmark     & \checkmark & \checkmark \\
 & Acceleration & \checkmark & \checkmark & \checkmark & \checkmark \\
\midrule
\multirow{2}{*}{Scene Context}
 & Object B.Boxes & \checkmark & \xmark & \checkmark & \xmark \\
 & Object Class   & \checkmark & \checkmark & \checkmark & \xmark \\
\midrule
Motion Captions & -- & \checkmark & \checkmark & \checkmark & \checkmark \\
\bottomrule
\end{tabularx}
\endgroup
\end{threeparttable}
\end{table}

\begin{table}[t]
\centering
\tiny
\begin{threeparttable}
\caption{Prompts for generating pedestrian motion captions. Each row describes one category and its corresponding prompt.}
\label{tab:motion_caps}
\begin{tabularx}{\linewidth}{lX}
\toprule
Category & Prompt \\
\midrule
Initial State and Readiness & What is the pedestrian doing at the start? Are they stationary, adjusting posture, shifting weight, or initiating movement? \\
Motion Trajectory and Dynamics & Is there consistent movement, acceleration, or deceleration? How does their position change over time? \\
Walking Direction & Are they walking perpendicular to the road (toward/across) or parallel (alongside)? Justify using their movement path and orientation. \\
Body Language and Orientation & Describe torso, leg, and arm alignment. Are they oriented toward the road or the crossing path? \\
Head and Gaze Behavior & Is their head facing towards the road, oncoming traffic, or scanning the scene, indicating awareness or intent to cross? \\
Interaction with Vehicles and Other Agents & Do they pause, slow down, or move in response to surrounding vehicles or pedestrians? \\
Signs of Hesitation or Yielding & Is there any visible hesitation or cautious behavior during the observation window? \\
Surface and Environment & What are they walking on? footpath, crosswalk, road? Do nearby elements support the idea of crossing intent? \\
Behavioral Transition & How does their behavior evolve across the N frames, e.g., from stationary to walking, or from observing to initiating a step? \\
Risk Awareness and Goal Inference & Are there signs of caution, urgency, or confidence that suggest a deliberate crossing decision (without guessing mental state)? \\
Future Behavior & Based on the visible behavior, what will the pedestrian most likely do in the next X seconds? \\
Scene Context & Are there road edges, traffic signals, or markings that help contextualize or influence the pedestrian's movement? \\
Final Position and Commitment & By the end of the observation window, has the pedestrian committed to crossing (e.g., stepped off curb, entered road)?\\
\bottomrule
\end{tabularx}
\end{threeparttable}
\end{table}

\subsection{Metadata Construction} 
We begin by collecting and consolidating all ground-truth, human-annotated labels provided by the constituent datasets, including pedestrian and interacting vehicle bounding boxes across frames, behavioral labels, ego-vehicle motion, and scene context (see Table~\ref{tab:dataset-annotations}). This metadata is normalized into a unified tabular schema to ensure consistency across datasets. 

\subsection{Pedestrian Motion Captioning}
To enrich the annotations with finer behavioral cues, we generate supplementary captions using a state-of-the-art VLM~\cite{qwen2.5vl} prompted with carefully designed pedestrian-motion prompts. These prompts are tailored to elicit fine-grained motion descriptions (see Table~\ref{tab:motion_caps}). Only the observation frames are used for this purpose. Each frame is preprocessed by extracting a $448 \times 448$ crop centered on the pedestrian and overlaying a red bounding box around the target individual across all crops, ensuring that the model consistently attends to the correct pedestrian and their immediate surroundings. Empirically, we found that feeding crops instead of full frames elicits more accurate responses from the VLM. The resulting captions are integrated with the original metadata.

\subsection{Prompt Design and Quality Assurance}
The metadata is transformed into a structured TSV format, where the rows correspond to frame indices and columns correspond to various attributes. Next, we construct a comprehensive prompt for LLM-based QA generation. In the prompt, we additionally incorporate a ``compliance checklist" to enforce rules for rationale generation, requiring the LLM to satisfy specific constraints for high-quality outputs and to regenerate whenever any constraint is violated. This checklist ensures that the rationales reflect information from all ground-truth annotation sources and motion captions. It also ensures that the rationales are not just \emph{parroted} attributes present in the metadata. 

The composite prompt now includes: (i) a system prompt establishing the task domain and constraints, (ii) explicit definitions of PIP and PTP tasks, (iii) detailed instructions for producing structured rationales across five reasoning categories (spatial, temporal, mathematical, ego-vehicle, and scene context), a final destinaton prediction, and a simple conclusion for everyday users, (iv) a curated set of in-context examples to guide reasoning style and output structure, (v) the compliance checklist, and (vi) the full sequence metadata. 

This prompt is fed into the \texttt{claude-sonnet-4} LLM-API~\cite{claude_sonnet_4}, which outputs question–answer–rationale triplets for each pedestrian sequence. Our pipeline yields a scalable method for producing multimodal reasoning annotations.  We record $10,251$ samples in our training set and $4,059$ in our test set.

\subsection{Rationale Generation}
To ensure that intention and trajectory predictions are explainable, PedestrianQA requires models to generate structured rationales spanning multiple scene dimensions. The rationales are categorized and described as follows:
\begin{enumerate}
    \item \textbf{Spatial Reasoning} captures the pedestrian’s physical state in the environment, including their pose, body orientation (e.g., parallel or perpendicular to the road), and exact spatial placement (e.g., standing on a curb or walking along a lane).
    \item \textbf{Temporal Reasoning} accounts for how motion evolves over time, such as the initiation of walking, acceleration or deceleration, and pauses before crossing. This category of rationale refers to frame numbers as timestamps.
    \item \textbf{Mathematical Reasoning} introduces a quantitative perspective, incorporating pedestrian–vehicle distance, trajectory angle, pedestrian velocity estimates, and displacement.
    \item \textbf{Ego-Vehicle Reasoning} links the pedestrian’s decisions to the behavior of the ego-vehicle: changes in speed, acceleration/deceleration, and braking that may enable or discourage crossing.
    \item \textbf{Scene-Context Reasoning} situates the pedestrian within a broader environment, drawing on cues such as traffic lights, crosswalks, road infrastructure, illumination, and interactions with other agents.
    \item \textbf{Final Destination Prediction} requires the model to infer the pedestrian’s likely endpoint within the scene, at the end of their trajectory. For example, reaching the opposite curb, halting midway, or continuing along the same side of the road.
    \item \textbf{Conclusion} summarizes the overall judgment in simple, accessible terms, providing a binary decision for intention prediction (e.g., cross or not cross) or a concise description of the forecasted trajectory.
\end{enumerate}

\section{Baseline Model}
We finetune the \texttt{Qwen2.5-VL-3B-Instruct} model~\cite{qwen2.5vl} on the PedestrianQA dataset using the official implementation\footnote{\url{https://github.com/QwenLM/Qwen2.5-VL/}}. Both PIP and PTP tasks are framed as QAs, with training data drawn from the official train splits (excluding validation) of all constituent datasets. We instruction-finetune the model using parameter-efficient LoRA~\cite{hu2022lora} adapters ($rank=8$, $\alpha=16$) to prevent overfitting. Empirically, we observe consistent gains on intention classification, trajectory prediction, and reasoning tasks, suggesting that the dataset size is sufficient for targeted capability adaptation. For JAAD~\cite{JAAD}, PIE~\cite{PIE}, and IDD-PeD~\cite{idd2025ped}, we finetune the model with $15$ input frames (corresponding to $0.5s$ of observation time), whereas, for TITAN~\cite{TITAN}, we use $10$ input frames (corresponding to $1s$ of observation time). All finetuning runs are performed on the original video resolution, with default hyperparameters, for 3 epochs on \texttt{2$\times$Nvidia RTX-A6000} GPUs.


\begin{table*}[t]
\caption{\textbf{PIP Results.} Performance comparison of \colorbox{green!10}{\textbf{Driving-specific VLM}}, \colorbox{cyan!10}{\textbf{Large-scale General-Purpose VLMs}},
\colorbox{yellow!20}{\textbf{Small-scale General-Purpose VLMs}},
 and our finetuned \colorbox{OrangeRed!15}{\textbf{Qwen2.5-VL-3b}}~\cite{qwen2.5vl} models. Finetune indicates whether the model was finetuned (\ding{51}) or evaluated in a zero-shot (\ding{55}) setting. \textbf{Bold} and \underline{underline} is for best and second-best results.}
\centering
\scriptsize   
\setlength{\tabcolsep}{4pt} 
\renewcommand{\arraystretch}{1.05}
\begin{tabular*}{\textwidth}{@{\extracolsep{\fill}}l c c c c c c c c c c c c}
\toprule
\textbf{Model} & \textbf{Params} & \textbf{Finetune}
& \multicolumn{2}{c}{\textbf{PIE~\cite{PIE}}}
& \multicolumn{2}{c}{\textbf{JAAD~\cite{JAAD}}}
& \multicolumn{2}{c}{\textbf{TITAN~\cite{TITAN}}}
& \multicolumn{2}{c}{\textbf{IDD-PeD~\cite{idd2025ped}}}
& \multicolumn{2}{c}{\textbf{Overall}} \\
\cmidrule(lr){4-5}\cmidrule(lr){6-7}\cmidrule(lr){8-9}\cmidrule(lr){10-11}\cmidrule(lr){12-13}
& & & Acc. & F$_1$ & Acc. & F$_1$ & Acc. & F$_1$ & Acc. & F$_1$ & Acc. & F$_1$ \\
\midrule
\rowcolor{green!10} Dolphins-9B~\cite{dolphins} & 9B & \ding{55} & 0.667 & 0.800 & 0.554 & 0.713 & 0.072 & 0.134 & 0.084 & 0.155 & 0.239 & 0.386 \\
\rowcolor{cyan!10} InternVL3-8B-Instruct~\cite{internvl3} & 8B & \ding{55} & 0.466 & 0.451 & 0.509 & 0.460 & 0.749 & 0.130 & 0.663 & 0.130 & 0.635 & 0.298 \\
\rowcolor{cyan!10} Kwai-Keye-8B~\cite{kwai-keye} & 8B & \ding{55} & 0.603 & 0.723 & 0.576 & 0.704 & 0.226 & 0.127 & 0.297 & 0.144 & 0.361 & 0.370 \\
\rowcolor{cyan!10} LLaVA-NeXT-Video-7B~\cite{llavanextvideo} & 7B & \ding{55} & 0.411 & 0.301 & 0.446 & 0.039 & 0.578 & 0.127 & \textbf{0.910} & 0.032 & 0.675 & 0.193 \\
\rowcolor{cyan!10} Qwen2.5-VL-7B-Instruct~\cite{qwen2.5vl} & 7B & \ding{55} & 0.449 & 0.323 & \textbf{0.627} & 0.548 & \textbf{0.926} & 0.029 & 0.876 & 0.087 & \underline{0.780} & 0.291 \\
\rowcolor{yellow!20} Qwen2.5-VL-3B-Instruct~\cite{qwen2.5vl} & 3B & \ding{55} & 0.667 & 0.800 & 0.554 & \underline{0.713} & 0.072 & 0.134 & 0.084 & 0.155 & 0.239 & 0.386 \\
\rowcolor{yellow!20} InternVL3-2B-Instruct~\cite{internvl3} & 2B & \ding{55} & 0.557 & 0.635 & 0.475 & 0.513 & 0.531 & 0.145 & 0.488 & 0.137 & 0.515 & 0.352 \\
\rowcolor{OrangeRed!15} \textbf{Ours (PIE)} & 3B & \ding{51} & \underline{0.667} & \textbf{0.800} & 0.554 & 0.705 & 0.073 & 0.134 & 0.102 & 0.158 & 0.247 & 0.389 \\
\rowcolor{OrangeRed!15} \textbf{Ours (JAAD)} & 3B & \ding{51} & 0.667 & \underline{0.795} & 0.554 & \textbf{0.713} & 0.072 & 0.134 & 0.084 & 0.155 & 0.239 & 0.386 \\
\rowcolor{OrangeRed!15} \textbf{Ours (TITAN)} & 3B & \ding{51} & \textbf{0.670} & 0.792 & 0.576 & 0.664 & 0.673 & 0.200 & 0.572 & 0.188 & 0.624 & 0.514 \\
\rowcolor{OrangeRed!15} \textbf{Ours (IDD-PeD)} & 3B & \ding{51} & 0.665 & 0.754 & \underline{0.588} & 0.633 & 0.733 & \textbf{0.203} & 0.867 & \underline{0.204} & 0.767 & \textbf{0.568} \\
\rowcolor{OrangeRed!15} \textbf{Ours (All Datasets)} & 3B & \ding{51} & 0.633 & 0.709 & 0.531 & 0.497 & \underline{0.808} & \underline{0.201} & \underline{0.880} & \textbf{0.205} & \textbf{0.783} & \underline{0.542} \\
\bottomrule
\newline
\newline
\end{tabular*}
\label{tab:pip_benchmark}
\vspace{-0.5cm}
\end{table*}

\begin{table*}[t]
\caption{\textbf{Performance comparison of PTP results.} \ding{51} indicates finetuned models, \ding{55} zero-shot. \textbf{Bold} marks best (lowest) values, \underline{underline} second-best. Dolphins~\cite{dolphins} was excluded from evaluation as it cannot generate trajectory bounding boxes required for this analysis. ADE and FDE are measured in pixels, in the image coordinate space.}
\centering
\scriptsize
\setlength{\tabcolsep}{4pt}
\renewcommand{\arraystretch}{1.05}
\begin{tabular*}{\textwidth}{@{\extracolsep{\fill}}l c c
    c c  c c  c c  c c  c c}
\toprule
\textbf{Model} & \textbf{Params} & \textbf{Finetune}
& \multicolumn{2}{c}{\textbf{PIE~\cite{PIE}}}
& \multicolumn{2}{c}{\textbf{JAAD~\cite{JAAD}}}
& \multicolumn{2}{c}{\textbf{TITAN~\cite{TITAN}}}
& \multicolumn{2}{c}{\textbf{IDD-PeD~\cite{idd2025ped}}}
& \multicolumn{2}{c}{\textbf{Overall}} \\
\cmidrule(lr){4-5}\cmidrule(lr){6-7}\cmidrule(lr){8-9}\cmidrule(lr){10-11}\cmidrule(lr){12-13}
& & & ADE & FDE & ADE & FDE & ADE & FDE & ADE & FDE & ADE & FDE \\
\midrule
\rowcolor{cyan!10} InternVL3-8B-Instruct~\cite{internvl3} & 8B & \ding{55} & 5937 & 19831 & 6118 & 15619 & 657 & 782 & 279 & 302 & 28414 & 8052 \\
\rowcolor{cyan!10} Kwai-Keye-8B~\cite{kwai-keye} & 8B & \ding{55} & 54 & 104 & 61 & 124 & \textbf{30} & \textbf{56} & 66 & 123 & 44 & 85 \\
\rowcolor{cyan!10} LLaVA-NeXT-Video-7B~\cite{llavanextvideo} & 7B & \ding{55} & 190 & 350 & 143 & 204 & 135 & 243 & 141 & 158 & 1522 & 261 \\
\rowcolor{cyan!10} Qwen2.5-VL-7B-Instruct~\cite{qwen2.5vl} & 7B & \ding{55} & \textbf{33} & \textbf{52} & 42 & \textbf{84} & 37 & \underline{57} & 56 & 98 & \underline{38} & \textbf{63} \\
\rowcolor{yellow!20} Qwen2.5-VL-3B-Instruct~\cite{qwen2.5vl} & 3B & \ding{55} & 96 & 176 & 54 & 101 & 43 & 66 & 61 & 106 & 61 & 105 \\
\rowcolor{yellow!20} InternVL3-2B-Instruct~\cite{internvl3} & 2B & \ding{55} & 827 & 1552 & 435 & 751 & 293 & 464 & 350 & 586 & 4632 & 811 \\
\rowcolor{OrangeRed!15} \textbf{Ours (PIE)} & 3B & \ding{51} & 51 & 87 & \underline{43} & 89 & 36 & 58 & 54 & 94 & 43 & 74 \\
\rowcolor{OrangeRed!15} \textbf{Ours (JAAD)} & 3B & \ding{51} & 93 & 167 & 56 & 105 & 43 & 69 & 60 & 104 & 60 & 104 \\
\rowcolor{OrangeRed!15} \textbf{Ours (TITAN)} & 3B & \ding{51} & 55 & 94 & 43 & 89 & 35 & 64 & \underline{49} & \underline{87} & \underline{43} & 78 \\
\rowcolor{OrangeRed!15} \textbf{Ours (IDD-PeD)} & 3B & \ding{51} & 71 & 127 & 44 & 88 & 37 & 58 & 50 & 88 & 49 & 84 \\
\rowcolor{OrangeRed!15} \textbf{Ours (All Datasets)} & 3B & \ding{51} & \underline{41} & \underline{70} & \textbf{41} & \underline{87} & \underline{31} & 60 & \textbf{46} & \textbf{80} & \textbf{37} & \underline{68} \\
\bottomrule
\newline
\newline
\end{tabular*}
\label{tab:ptp_benchmark}
\vspace{-0.5cm}
\end{table*}

\begin{table*}[t]
\caption{\textbf{Rationale evaluation on the \textbf{combined dataset}, with Claude-Sonnet-4.} Average scores (0--100) for Spatial Reasoning (SR), Temporal Reasoning (TR), Mathematical Reasoning (MR), Ego-Vehicle Reasoning (EVR), Scene-Context Reasoning (SCR), Final Destination Prediction (FDP), and Conclusion (C). \ding{51} indicates finetuned models, \ding{55} zero-shot. \textbf{Bold} shows best score per column; \underline{underline} marks the second-best. Dolphins~\cite{dolphins} generates only a brief conclusion and does not generate category-specific rationales.}
\centering
\scriptsize
\setlength{\tabcolsep}{4pt}
\renewcommand{\arraystretch}{1.05}
\begin{tabular*}{\textwidth}{@{\extracolsep{\fill}}l c c
    c c c c c c c}
\toprule
\textbf{Model} & \textbf{Params} & \textbf{Finetune}
& \textbf{SR} & \textbf{TR} & \textbf{MR} & \textbf{EVR} & \textbf{SCR} & \textbf{FDP} & \textbf{C} \\
\midrule
\rowcolor{green!10} Dolphins-9B~\cite{dolphins} & 9B & \ding{55} & 0.00 & 0.00 & 0.00 & 0.00 & 0.00 & 0.00 & 12.05 \\
\rowcolor{cyan!10} InternVL3-8B-Instruct~\cite{internvl3} & 8B & \ding{55} & 37.01 & 37.60 & 26.36 & 29.17 & 37.79 & 29.63 & 45.05 \\
\rowcolor{cyan!10} Kwai-Keye-8B~\cite{kwai-keye} & 8B & \ding{55} & 33.15 & 30.74 & 28.10 & 36.38 & 37.09 & \textbf{38.03} & 28.70 \\
\rowcolor{cyan!10} LLaVA-NeXT-Video-7B~\cite{llavanextvideo} & 7B & \ding{55} & 46.39 & 33.96 & 39.52 & 40.06 & 28.77 & 17.22 & 50.33 \\
\rowcolor{cyan!10} Qwen2.5-VL-7B-Instruct~\cite{qwen2.5vl} & 7B & \ding{55} & 40.15 & 41.90 & 31.74 & 38.08 & 38.04 & 12.76 & 50.89 \\
\rowcolor{yellow!20} Qwen2.5-VL-3B-Instruct~\cite{qwen2.5vl} & 3B & \ding{55} & 22.05 & 19.92 & 17.25 & 27.11 & 28.80 & 10.48 & 14.80 \\
\rowcolor{yellow!20} InternVL3-2B-Instruct~\cite{internvl3} & 2B & \ding{55} & 31.71 & 32.30 & 23.02 & 27.47 & 34.25 & 25.95 & 37.09 \\
\rowcolor{OrangeRed!15} \textbf{Ours (PIE)} & 3B & \ding{51} & 25.52 & 24.48 & 22.00 & 36.70 & 30.43 & 14.58 & 18.28 \\
\rowcolor{OrangeRed!15} \textbf{Ours (JAAD)} & 3B & \ding{51} & 21.85 & 19.84 & 17.03 & 26.99 & 28.46 & 10.95 & 15.00 \\
\rowcolor{OrangeRed!15} \textbf{Ours (TITAN)} & 3B & \ding{51} & 46.26 & 41.71 & 39.77 & 45.25 & 47.38 & 22.80 & 43.06 \\
\rowcolor{OrangeRed!15} \textbf{Ours (IDD-PeD)} & 3B & \ding{51} & \underline{56.33} & \underline{51.35} & \underline{46.83} & \underline{57.77} & \underline{55.63} & 26.53 & \underline{55.37} \\
\rowcolor{OrangeRed!15} \textbf{Ours (All Datasets)} & 3B & \ding{51} & \textbf{58.36} & \textbf{54.84} & \textbf{51.68} & \textbf{61.51} & \textbf{59.72} & \underline{32.24} & \textbf{60.25} \\
\bottomrule
\end{tabular*}
\label{tab:rationale_benchmark}
\vspace{-0.5cm}
\end{table*}
\section{Experiments}
\label{benchmark-sec}

\subsection{Experimental Settings}

Our benchmark includes a representative set of widely used VLMs: Dolphins~\cite{dolphins}, LLaVA-NeXt~\cite{llavanextvideo}, and InternVL3~\cite{internvl3}, Qwen2.5-VL~\cite{qwen2.5vl}, together with Kwai-Keye~\cite{kwai-keye}, a more recent rationale-generation model. Each model is prompted with task-specific instructions tailored to itself, and optimized for prediction and rationale generation, for both PIP and PTP tasks. We evaluate on the default test sets of all constituent datasets. For PIP, we formulate the task as binary classification, report accuracy and $F_1$-score. For PTP, we compute trajectory forecasting metrics: Average Displacement Error (ADE) and Final Displacement Error (FDE). To assess caption quality, we adapt the CLAIR framework~\cite{clair} to our domain and employ Claude-Sonnet-4~\cite{claude_sonnet_4} as the evaluator.

\subsection{Pedestrian Intention Prediction Results}
In Table~\ref{tab:pip_benchmark}, we observe that \texttt{Ours (All Datasets)} model achieves the highest overall accuracy of 78.3\%, narrowly outperforming a much larger \texttt{Qwen2.5-VL-7B-Instruct}, and comfortably outperforming other 2b/3b-parameter VLMs. These include \texttt{Qwen2.5-VL-3B-Instruct} and \texttt{InternVL3-2B-Instruct}, which our finetuned model outperforms by 54.4\% and 26.8\% on overall accuracy, respectively. We also highlight that while \texttt{Qwen2.5-VL-7B-Instruct} achieves impressive performance, our finetuned model outperforms other models of similar scale: \texttt{Dolphins-9B}, \texttt{InternVL3-8B-Instruct}, \texttt{Kwai-Keye-8B}, and \texttt{LLaVA-NeXT-Video-7B} by 54.4\%, 14.8\%, 42.2\%, and 10.8\%, respectively. We attribute the performance gains to the finetuned model’s ability to recognize subtle cues of crossing behavior before the action occurs. The model attends to pedestrian orientation, pose, and surrounding traffic in structured environments, while the unstructured datasets contribute edge-case examples of erratic pedestrian behavior. This effect is evident in our IDD-PeD-only and TITAN-only fine-tuned models, which achieve strong results on the PIE and JAAD datasets. The high accuracy and low F1-score stem from the strong class imbalance between positive and negative crossing samples, with the negative class being far more prevalent. \texttt{Ours (All Datasets)} distinguishes the classes more effectively, achieving an overall F1-score that is 25.1\% higher than that of \texttt{Qwen2.5-VL-7B-Instruct}.

\subsection{Pedestrian Trajectory Prediction Results}
In Table~\ref{tab:ptp_benchmark}, we observe that our finetuned model \texttt{Ours (All Datasets)} achieves the lowest overall ADE, while narrowly losing out to a much larger \texttt{Qwen2.5-VL-7B} model in the FDE metric, by just 5 pixels. The same finetuned model achieves the best ADE in JAAD and IDD-PeD. Our model loses out to \texttt{Qwen2.5-VL-7B} in the PIE test set by just 9 pixels. Once again, we observe that our finetuned model \texttt{Ours (All Datasets)} achieves a much lower displacement error compared to other models of similar size: \texttt{InternVL3-2B} by 426 pixels ADE and 743 pixels FDE, and \texttt{Qwen2.5-VL-3B} by 24 pixels ADE and 37 pixels FDE on the overall result. This performance gain arises from jointly training the model on PIP and PTP tasks, enhancing its ability to infer crossing intentions accurately. Pedestrians oriented perpendicular to the road are considerably more likely to cross and exhibit trajectories distinct from those parallel to it. We notice that \texttt{Kwai-Keye-8B} achieves the lowest error on the TITAN dataset, narrowly beating our \texttt{Ours (All Datasets)} by just 1 pixel ADE and 4 pixels FDE. The rest of our finetuned models remain comparable to \texttt{Qwen2.5-VL-7B} and \texttt{Kwai-Keye-8B}.

\subsection{Rationale Generation Results} 
In rationale evaluation (Table~\ref{tab:rationale_benchmark}), on the combined dataset, our finetuned \texttt{Ours (All Datasets)} achieves the highest score in 6 out of 7 types of captions, losing out to \texttt{Kwai-Keye-8B} in the Final Destination Prediction category by nearly 6\%. In other categories, it outperforms \texttt{Qwen2.5-VL-7B} by as much as 23.43\% (Ego-Vehicle Reasoning). While we use \texttt{claude-sonnet-4} for both generation and evaluation, scores remain well below saturation $(\leq62\%)$. The evaluation process penalizes hallucinations and weak visual grounding, thus preventing circular bias.

\subsection{Qualitative Analysis}
\begin{figure}[t]
  \centering
  \includegraphics[width=\columnwidth]{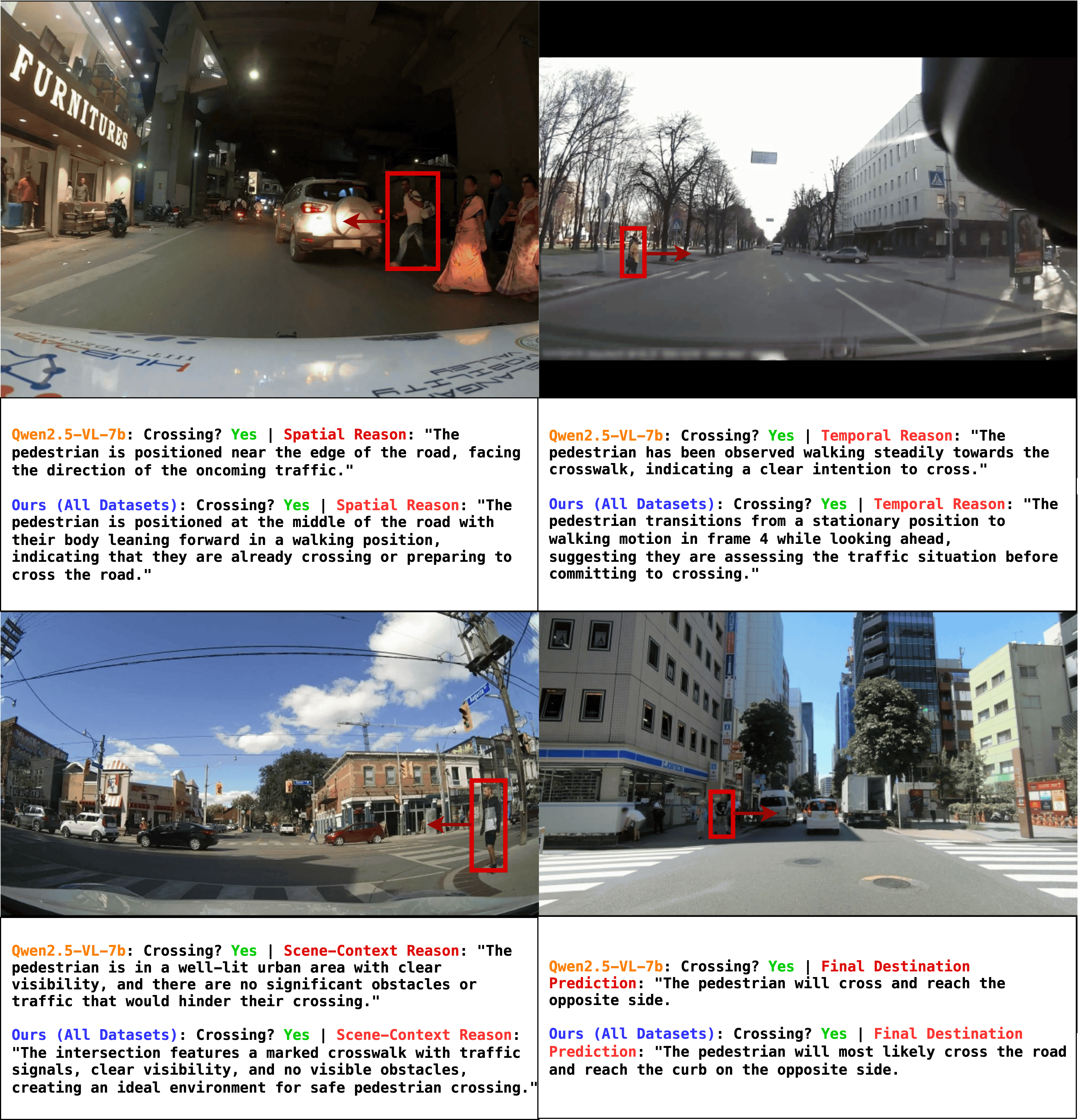}
  \caption{Qualitative analysis between \texttt{Qwen2.5-VL-7b} and \texttt{Ours (All Datasets)} on the compositional datasets (clockwise: IDD-PeD~\cite{idd2025ped}, JAAD~\cite{JAAD}, TITAN~\cite{TITAN}, PIE~\cite{PIE}). We identify samples having correct predicate predictions for a fair comparison among rationales.}
  \label{fig:qual_analysis}
\end{figure}

We establish a qualitative analysis between our strongest model: \texttt{Ours (All Datasets)} and the strongest default \texttt{Qwen-2.5-VL-7b} model in Fig.~\ref{fig:qual_analysis}. Our model generates rationales with more accurate object grounding (e.g., detecting crosswalks and traffic lights in PIE) and richer context from visual cues (e.g., identifying the curb as the destination in TITAN). The rationales are also more detailed and less prone to hallucinations. For example, in the IDD-PeD example, our model correctly recognizes a pedestrian standing in the middle of the road. Whereas the \texttt{Qwen2.5-VL-7b} baseline incorrectly assumes the starting position is on the edge of the road due to its higher statistical bias of a positive crossing intention.

\subsection{Effect of Training on Unstructured Traffic Data} 
Interestingly, our model trained only on IDD-PeD~\cite{idd2025ped} maintains a comparable performance with our model trained on all datasets (1 to 5\% lower rationale scores). This pattern can also be noticed in PTP tasks in the JAAD-ADE (44 vs 41 pixels) and the TITAN-ADE (37 vs 31 pixels) metrics. In PIP, this behavior is observed in PIE accuracy (higher at 66.5\% compared to 63.3\%), JAAD accuracy (higher at 58.8\% compared to 53.1\%), comparable performance in TITAN accuracy, and narrowly losing out by 1.6\% in the overall accuracy. Based on these observations, we conclude that sufficiently powerful models like ~\cite{qwen2.5vl}, if trained only on IDD-PeD~\cite{idd2025ped}, a dataset containing only chaotic unstructured traffic scenarios and a much higher density of traffic agents, can have comparable or better performance over similar models trained on structured datasets.

\section{Conclusion}
This work reframes pedestrian behavior understanding as a multimodal reasoning problem. It introduces PedestrianQA, a video-based QA corpus that jointly evaluates: (i) PIP, (ii) PTP, and (iii) fine-grained, object-grounded spatio-temporal rationales. By coupling short-horizon visual evidence with structured explanations, PedestrianQA encourages models to predict \emph{what} will happen and to justify \emph{why} it is plausible, advancing explainability for safety-critical autonomous driving. Empirically, finetuning state-of-the-art VLMs on PedestrianQA yields consistent gains across structured (PIE~\cite{PIE}, JAAD~\cite{JAAD}, TITAN~\cite{TITAN}), and unstructured (IDD-PeD~\cite{idd2025ped}) datasets, improving both decision accuracy (PIP) and motion forecasting quality (PTP), while producing more informative and category-aligned rationales. Notably, compact models finetuned on our corpus approach or surpass the performance of substantially larger baselines, suggesting that task-specific multimodal supervision is a powerful complement to internet-scale pretraining. The strong transfer observed from IDD-PeD-only training to other datasets indicates that dense, unstructured scenarios provide rich supervisory signals for generalization.
\newline
\textbf{Acknowledgment.} This project was supported by iHub-Data and Mobility at IIIT Hyderabad. 


\end{document}